\documentclass[10pt, conference]{IEEEtran} 
\IEEEoverridecommandlockouts
\usepackage{amsmath,amssymb,amsfonts}
\usepackage{algorithmic}
\usepackage{graphicx}
\usepackage{textcomp}
\usepackage{xcolor}
\usepackage{ragged2e} 
\usepackage{hyperref}
\usepackage{float}
\usepackage{multirow}
\usepackage{booktabs}
\usepackage{listings}
\usepackage{xcolor}
\usepackage{bm}  
\usepackage{xspace}

\begin{document}
\title{\Large \bfseries Enhancing Financial VQA in Vision Language Models using Intermediate Structured Representations }
\author{
  \begin{minipage}{0.5\textwidth}
    \centering
    \textbf{Archita Srivastava} \\
    Data Scientist \\
    Synechron, Bangalore, India\\
    \texttt{archita.srivastava1@synechron.com} 
  \end{minipage}%
  \hspace{0.001\textwidth}%
  \begin{minipage}{0.5\textwidth}
    \centering
    \textbf{Abhas Kumar} \\
    Lead Data Scientist \\
    Synechron, Bangalore, India\\
    \texttt{abhas.kumar@synechron.com}
  \end{minipage} \\[3em]  
  \begin{minipage}{0.5\textwidth}
    \centering
    \textbf{Rajesh Kumar} \\
    Lead Data Scientist \\
    Synechron, Bangalore, India\\
    \texttt{rajesh.kumar3@synechron.com} 
  \end{minipage}%
  \hspace{0.001\textwidth}%
  \begin{minipage}{0.5\textwidth}
    \centering
    \textbf{Prabhakar Srinivasan} \\
    Director  \\
    Synechron, Bangalore, India\\
    \texttt{prabhakar.srinivasan@synechron.com}
  \end{minipage}
}

\maketitle

\centering
\section*{\textbf{Abstract}}
\justify
Chart interpretation is crucial for visual data analysis, but accurately extracting information from charts poses significant challenges for automated models. This study investigates the fine-tuning of DEPLOT, a modality conversion module that translates the image of a plot or chart to a linearized table, on a custom dataset of 50,000 bar charts. The dataset comprises simple, stacked, and grouped bar charts, targeting the unique structural features of these visualizations. The fine-tuned DEPLOT model is evaluated against its base version using a test set of 1,000 images and two metrics: Relative Mapping Similarity (RMS), which measures categorical mapping accuracy, and Relative Number Set Similarity (RNSS), which evaluates numerical interpretation accuracy.
To further explore the reasoning capabilities of large language models (LLMs), we curate an additional set of 100 bar chart images paired with question-answer sets. Our findings demonstrate that providing a structured intermediate table alongside the image significantly enhances LLM reasoning performance compared to direct image queries.

\section*{\textbf{1. Introduction}}
\justify
In today’s data-driven world, data visualization plays a crucial role in conveying complex information clearly and effectively. Charts, such as bar charts, line charts, and pie charts, are fundamental tools for this purpose, widely applied across domains like finance, healthcare, business intelligence, and scientific research to support decision-making. However, accurately extracting and interpreting information from charts remains a significant challenge for automated models due to the wide variation in chart structures, configurations, and complexities. This difficulty is especially pronounced in domain-specific applications, where visualizations often possess unique characteristics tailored to specialized data needs.

DEPLOT \cite{liu2023DEPLOToneshotvisuallanguage} introduces a modality conversion module that maps visual chart data into structured data tables, effectively translating visual features into data representations. Although DEPLOT is trained on diverse chart types, it can achieve enhanced accuracy through fine-tuning on domain-specific datasets. This targeted approach enables the model to learn the distinctive features and attributes of specific chart types, improving both categorical mapping and numerical interpretation. Once accurate intermediate tables are generated, large language models (LLMs) \cite{brown2020languagemodelsfewshotlearners} can leverage their few-shot learning capabilities for advanced reasoning and querying tasks.

To evaluate the effectiveness of the fine-tuned model, we employ two key metrics: Relative Number Set Similarity (RNSS) \cite{masry-etal-2022-chartqa, luo2021chartocr} and Relative Mapping Similarity (RMS). RNSS measures the relative accuracy of predicted numerical values compared to ground-truth data, tolerating minor transpositions and small errors. RMS assesses the accuracy of categorical label mappings, accounting for textual and numerical distances between predicted and actual entries. By calculating RMS precision, recall, and F1 scores, we comprehensively evaluate the model’s ability to interpret table structures while remaining invariant to row and column permutations.

Our results demonstrate that fine-tuning DEPLOT on a custom dataset significantly improves its ability to accurately interpret both categorical and numerical aspects of bar chart data. This approach establishes a framework for adapting DEPLOT to various domains by tailoring datasets to specific chart types or configurations, paving the way for the development of highly accurate, domain-specific models for data visualization analysis. By highlighting the benefits of targeted fine-tuning, this research advances the field of visual-language models for automated chart interpretation, enabling more reliable and adaptable data extraction from visualizations.

\section*{\textbf{2. Problem Statement}}
\justify
Charts are a cornerstone of data visualization, widely used to represent categorical data and facilitate comparisons across groups. However, their structural diversity and domain-specific configurations pose significant challenges for automated interpretation. Current systems struggle to extract and interpret information accurately due to varying chart designs, overlapping visual elements, and complex layouts.

Hybrid methods relying on OCR and rule-based systems are often limited by their rigidity and dependence on manual configurations, making them unsuitable for handling diverse chart structures. Meanwhile, end-to-end models, though promising, demand extensive fine-tuning on large datasets and still exhibit shortcomings when addressing complex queries or unseen configurations. In domains like finance and scientific research, where precision and contextual understanding are critical, these limitations hinder the deployment of reliable automated solutions.

The need for a robust, scalable approach capable of handling diverse chart types and configurations, while ensuring precise numerical and categorical data extraction, remains unmet. Addressing this gap requires integrating structured intermediate representations and leveraging advanced reasoning models to enhance performance in interpreting and querying bar charts.

\section*{\textbf{3. Prior Methodologies}}
\justify
The field of automated chart interpretation has seen two primary methodological approaches: hybrid systems and end-to-end models. Hybrid systems combine optical character recognition (OCR), keypoint detection, and object segmentation with hand-engineered rules to extract chart data \cite{10.1007/978-3-319-46478-7_41, luo2021chartocr, masry-etal-2022-chartqa}. While effective for simple scenarios, these systems face scalability challenges due to their reliance on domain-specific rules and manual configuration.

End-to-end models, on the other hand, bypass intermediate steps by directly interpreting visual chart data \cite{liu2023matchaenhancingvisuallanguage} \cite{lee2023pix2structscreenshotparsingpretraining}. Although promising, these methods demand extensive fine-tuning on specialized datasets, limiting their adaptability and performance in handling complex or novel chart configurations. 

DEPLOT \cite{liu2023DEPLOToneshotvisuallanguage}, a modality conversion module, bridges the gap by translating visual chart data into structured tables. This intermediate representation enables downstream models, such as large language models (LLMs), to perform reasoning tasks with enhanced accuracy. Despite its versatility, the base DEPLOT model struggles with domain-specific charts, necessitating fine-tuning to improve its performance for specific applications. 

\section*{\textbf{4. Our Proposed Solution}}
\justify
To address the limitations of existing methods, we propose fine-tuning the DEPLOT model on a custom dataset of 50,000 bar charts, including simple, stacked, and grouped configurations. This targeted fine-tuning allows the model to better capture the unique structural features of bar charts, enhancing its ability to generate accurate categorical mappings and numerical interpretations. By leveraging intermediate structured table representations generated by the fine-tuned DEPLOT model, large language models (LLMs) can more effectively reason and answer questions about chart data.

Our approach includes evaluating the fine-tuned model using two metrics: Relative Number Set Similarity (RNSS) and Relative Mapping Similarity (RMS). RNSS assesses numerical accuracy by comparing predicted values to ground-truth data, while RMS measures the precision, recall, and F1 scores of categorical label mappings. These metrics ensure a comprehensive evaluation of the model’s interpretative capabilities.

In addition to enhancing DEPLOT’s performance, we analyze the reasoning abilities of LLMs when provided with intermediate structured tables. Three experimental scenarios are considered:
\begin{enumerate}
    \item Query and image only,
    \item Query, image, and a table generated by the fine-tuned DEPLOT model ($F_t$), and
    \item Query, image, and a table generated by the base DEPLOT model ($B_t$).
\end{enumerate}
Our findings demonstrate that accurate table representations significantly enhance LLM reasoning capabilities, with smaller models like Qwen2-VL-7B \cite{wang2024qwen2vlenhancingvisionlanguagemodels} outperforming larger counterparts such as GPT-4o when provided with high-quality structured data. This research highlights the transformative potential of integrating modality conversion and fine-tuning to advance automated chart interpretation and reasoning.

\section*{\textbf{5. Dataset Creation}}
\justify
The performance of machine learning models, particularly those designed for chart interpretation like DEPLOT, relies heavily on the quality, diversity, and relevance of the training and evaluation datasets \cite{he2019dataqualitymodelquality} \cite{budach2022effectsdataqualitymachine}. In this study, we introduce a dataset of 50,000 bar charts, meticulously designed to represent financial applications. This dataset encompasses a wide range of financial metrics and concepts, providing the contextual depth needed to fine-tune DEPLOT for financial chart analysis. While tailored for financial applications, the methodology and techniques used in creating this dataset are highly adaptable and can be extended to other domains or chart attributes with minimal modification.

\subsection*{\textbf{5.1 Dataset Characteristics}} 
\justify
To capture a wide range of potential scenarios, key elements of the bar charts were randomized using a curated list of domain-specific labels and values tailored to financial applications. The randomized elements include:

\begin{itemize} 
    \item \textbf{Titles}: A diverse array of financial chart titles was generated, such as “Quarterly Revenue Growth,” “Market Share by Region,” and “Profitability Comparison.” These titles were carefully curated to reflect common themes in financial reporting and analysis, ensuring realistic context during chart generation.
    
    \item \textbf{X-Axis Labels}: The x-axis labels, typically representing time or categories, included options like “Fiscal Quarter,” “Region,” and “Product Category.” These were selected from a predefined list of terms relevant to financial analysis, ensuring contextual alignment with the domain.
    
    \item \textbf{Y-Axis Labels}: The y-axis reflected financial metrics such as “Revenue (\$),” “Profit Margin (\%),” and “Market Capitalization (\$B).” Randomized selection from a comprehensive list ensured clarity and relevance to financial data representation.
    
    \item \textbf{Categories}: Categories, representing data groupings such as financial years, regions, or product lines, were randomly assigned. This diversity helps the model generalize across various financial contexts.
    
    \item \textbf{Bar Ranges}: Value ranges for the bars were randomly generated to reflect realistic financial variations, including both positive and negative values. This aspect accurately represents the dynamic nature of financial data.
    
    \item \textbf{Bar Types}: The dataset incorporated multiple chart types, including simple, stacked, and grouped bar charts. Each type offers unique visual representations, enriching the dataset’s diversity.
    
    \item \textbf{Bar Orientation}: Both horizontal and vertical bar orientations were included to ensure the model could interpret financial data regardless of layout.
    
    \item \textbf{Annotations}: Some charts featured numerical annotations directly on the bars, while others omitted them. This variability introduces an additional layer of complexity, enabling the model to learn from both annotated and non-annotated visual cues.
\end{itemize}

\justify By randomizing these elements from a well-structured list, we ensure that the dataset is not only diverse and comprehensive but also specifically tailored to reflect the nuances of financial data representation. This careful design enhances the model’s ability to understand and interpret a wide range of financial charts accurately.

\subsection*{\textbf{5.2 Generalizability to Other Domains}} 
\justify
While the current dataset is tailored for financial applications, the methodology developed in this study is highly adaptable and can be extended to create datasets for a wide range of industries and domains.

The core approach involves customizing chart elements such as titles, axis labels, categories, and annotations to align with domain-specific terminology and concepts. This ensures that the model interprets charts within their appropriate context, whether they pertain to financial data, healthcare metrics, or marketing performance.

For example, in the healthcare domain, chart titles could include “Patient Survival Rates by Treatment” or “Hospital Admissions by Disease Type,” with corresponding y-axis labels such as “Survival Rate \%” or “Number of Admissions.” The bars could represent data points like diseases, treatment types, or patient demographics, randomized to create a diverse and robust dataset. Similarly, in the marketing sector, potential titles might be “Ad Performance Over Time” or “Revenue Growth by Region,” with y-axis labels such as “Conversion Rate \%” or “Sales Revenue \$.”

This versatile methodology allows for the development of specialized datasets across domains, enabling the fine-tuning of DEPLOT or similar models for various real-world applications. This approach transforms DEPLOT into a specialized tool that enhances its chart interpretation capabilities, enabling it to address sector-specific tasks effectively.

\subsection*{\textbf{5.3 Synthetic Training Data Generation}}
\justify
The bar charts for this dataset were generated using Python’s Matplotlib \cite{4160265} and Seaborn \cite{2021JOSS....6.3021W} libraries, offering a wide range of customization options for creating diverse, visually complex charts. Each chart was saved as an image file (e.g., .png) with an accompanying structured JSON file that contains the textual description necessary for model training. This textual description serves as the ground truth that aligns with the visual content of the chart.

To maintain consistency with the output format of the base DEPLOT model, we structured the textual descriptions carefully. Below are examples of the textual descriptions for various chart types, reflecting different layouts and their corresponding labels.

\justify
\textbf{General Structure of the Textual Descriptions}
\begin{itemize}
    \item \textbf{TITLE:} The title of the chart, which varies according to the chart's context and the domain.
    \item \textbf{X-Axis Label:} Denotes the category or dimension of the data, such as financial metrics, time periods, or regions. This label will change depending on the layout and type of data being represented (e.g., fiscal year, product category, or market region).
    \item \textbf{Y-Axis Label:} Represents the metric or value being measured, such as revenue, costs, or growth percentages. This label is adjusted according to the specific chart layout.
    \item \textbf{Categories and Values:} For each category on the x-axis, the corresponding value(s) for the y-axis are listed. Depending on the chart layout (simple, stacked, or grouped), the values are presented differently:
    \begin{itemize}
        \item For simple bar charts, each category will have a single value associated with it.
        \item For stacked bar charts, multiple values are stacked for each category to represent different sub-metrics.
        \item For grouped bar charts, multiple bars are grouped together for each category, each representing a different group or series.
    \end{itemize}
\end{itemize}

The ``\textbar " delimiter separates columns (i.e., labels and values), while ``\texttt{<0x0A>}" marks the end of each row, ensuring proper parsing and structure for the model to interpret the data accurately.

\subsection*{\textbf{Example Textual Descriptions for Different Chart Types}}

\begin{figure}[htbp]
    \centering
    \includegraphics[width=1\linewidth]{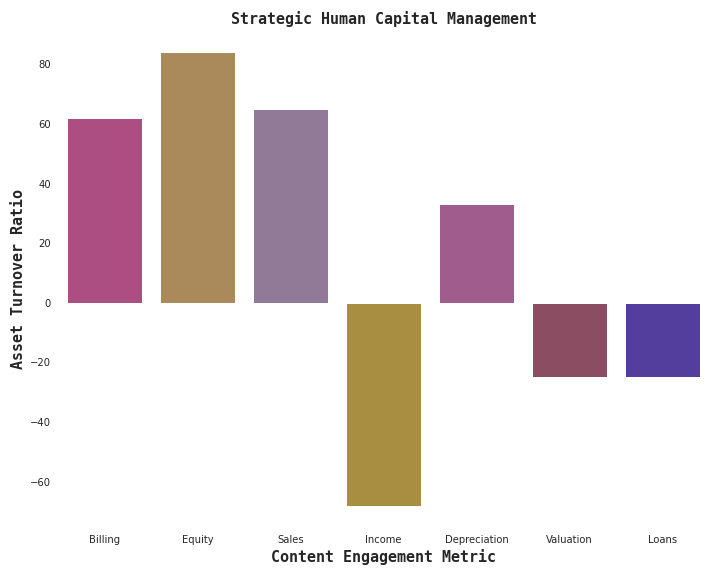}
    \caption{Simple Vertical Bar Chart}
    \label{fig: Simple Vertical Bar Chart}
\end{figure}

\justify \textbf{1. Simple Vertical Bar Chart:} An example chart, Fig. \ref{fig: Simple Vertical Bar Chart}, is provided to illustrate the syntax of a simple vertical bar chart along with its corresponding ground truth representation.  \\

\textit{``TITLE \textbar \{Chart Title\} \texttt{<0x0A>} \{X-Axis Label\} \textbar \{Y-Axis Label\} \texttt{<0x0A>} \{Category\_1\} \textbar \{Value\_1\} \texttt{<0x0A>} \{Category\_2\} \textbar \{Value\_2\} ..."} \\

\textbf{Example:} \\

\textit{``TITLE \textbar Strategic Human Capital Management \texttt{<0x0A>} Content Engagement Metric \textbar Asset Turnover Ratio \texttt{<0x0A>} Billing \textbar 62 \texttt{<0x0A>} Equity \textbar 84 \texttt{<0x0A>} Sales \textbar 65 \texttt{<0x0A>} Income \textbar -68 \texttt{<0x0A>} Depreciation \textbar 33 \texttt{<0x0A>} Valuation \textbar -25 \texttt{<0x0A>} Loans \textbar -25"}\\

\begin{figure}[htbp]
    \centering
    \includegraphics[width=1\linewidth]{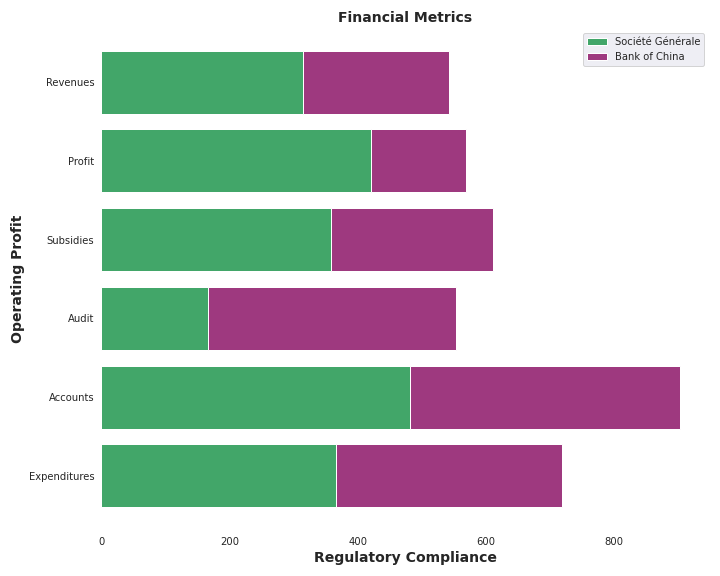}
    \caption{Stacked Horizontal Bar Chart}
    \label{fig: Stacked Horizontal Bar Chart}
\end{figure}

\justify \textbf{2. Stacked Horizontal Bar Chart:} An example chart, Fig. \ref{fig: Stacked Horizontal Bar Chart}, is provided to illustrate the syntax of a stacked horizontal bar chart along with its corresponding ground truth representation. \\

\textit{``TITLE \textbar \{Chart Title\} \texttt{<0x0A>} \{Y-Axis Label\} \textbar \{Stack\_A\} \textbar \{Stack\_B\} \texttt{<0x0A>} \{Category\_1\} \textbar \{Value\_1\_A\} \textbar \{Value\_1\_B\} \texttt{<0x0A>} \{Category\_2\} \textbar \{Value\_2\_A\} \textbar \{Value\_2\_B\} \ldots''}\\

\textbf{Example:} \\

\textit{``TITLE \textbar Financial Metrics \texttt{<0x0A>} Operating Profit \textbar Société Générale \textbar Bank of China \texttt{<0x0A>} Expenditures \textbar 366 \textbar 352 \texttt{<0x0A>} Accounts \textbar 482 \textbar 421 \texttt{<0x0A>} Audit \textbar 167 \textbar 386 \texttt{<0x0A>} Subsidies \textbar 358 \textbar 253 \texttt{<0x0A>} Profit \textbar 421 \textbar 147 \texttt{<0x0A>} Revenues \textbar 314 \textbar 228''}

\vspace{5mm}

\begin{figure}[htbp]
    \centering
    \includegraphics[width=1\linewidth]{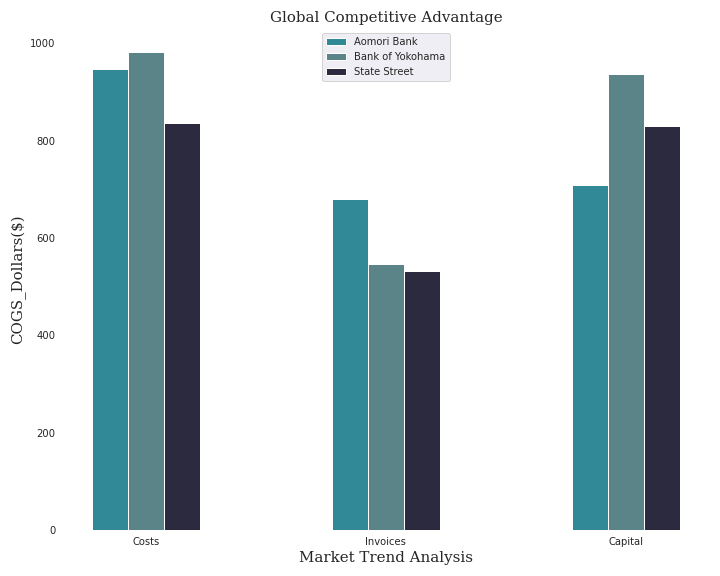}
    \caption{Grouped Vertical Bar Chart}
    \label{fig: Grouped Vertical Bar Chart}
\end{figure}

\justify \textbf{3. Grouped Vertical Bar Chart:} An example chart, Fig. \ref{fig: Grouped Vertical Bar Chart}, is provided to illustrate the syntax of a grouped vertical bar chart along with its corresponding ground truth representation.  \\

\textit{``TITLE \textbar \{Chart Title\} \texttt{<0x0A>} \{X-Axis Label\} \textbar \{Group\_A\} \textbar \{Group\_B\} \textbar \{Group\_C\} \texttt{<0x0A>} \{Category\_1\} \textbar \{Value\_1\_A\} \textbar \{Value\_1\_B\} \textbar \{Value\_1\_C\} \texttt{<0x0A>} \{Category\_2\} \textbar \{Value\_2\_A\} \textbar \{Value\_2\_B\} \textbar \{Value\_2\_C\} ..."} \\

\textbf{Example:} \\

\textit{``TITLE \textbar Global Competitive Advantage \texttt{<0x0A>} Market Trend Analysis \textbar Aomori Bank \textbar Bank of Yokohama \textbar State Street \texttt{<0x0A>} Costs \textbar 948 \textbar 983 \textbar 837 \texttt{<0x0A>} Invoices \textbar 680 \textbar 547 \textbar 532 \texttt{<0x0A>} Capital \textbar 709 \textbar 937 \textbar 830"}\\

\justify
Each description ensures that all chart types—whether simple, stacked, or grouped—are consistently formatted, with labels and categories adjusted according to the layout. The data is structured to support effective training of the DEPLOT model by providing a rich, diverse dataset that reflects various financial scenarios and chart configurations.

\section*{\textbf{6. Evaluation of Base DEPLOT Model on Custom Dataset}}
\justify
In this section, we evaluate the performance of the base DEPLOT model on 1,000 generated test images across three bar chart types—simple, stacked, and grouped. The model’s accuracy is assessed using two metrics: Relative Number Set Similarity (RNSS) and Relative Mapping Similarity (RMS), both of which offer different insights into the alignment between the predicted output and the target data. The RNSS metric focuses on numeric accuracy, while RMS provides a comprehensive evaluation by incorporating structural information and alignment between the row-column mappings.

\subsection*{\textbf{6.1 Metrics Overview}}
\justify

{\textbf{A. Relative Number Set Similarity (RNSS) : }} \
\justify
The RNSS metric \cite{masry-etal-2022-chartqa} is used to evaluate the similarity between the sets of numeric entries predicted by the model and those in the target table. This metric assesses how closely the set of predicted numbers aligns with the ground-truth values in the target table without considering the position of these values.

To compute RNSS, let \( P = \{ p_i \}_{1 \leq i \leq N} \) represent the set of model-predicted numbers in the table and \( T = \{ t_j \}_{1 \leq j \leq M} \) represent the numbers in the target table. A threshold of 10\% is applied to account for minor deviations, allowing predicted values \( p_i \) to be considered correct if they lie within \( \pm 10\% \) of the corresponding ground-truth values \( t_j \). 

\justify
The relative distance \( D(p, t) \) between each predicted and target number is calculated as:

\[
D(p, t) = \min\left(1, \frac{|p - t|}{|t|}\right)
\]

The final RNSS score is then computed by finding the minimal cost matching between elements in \( P \) and \( T \) using a binary matrix \( X \in \mathbb{R}^{N \times M} \). The score is defined as:

\[
\text{RNSS} = 1 - \frac{\sum_{i=1}^{N} \sum_{j=1}^{M} X_{ij} D(p_i, t_j)}{\max(N, M)}
\]

\justify

However, RNSS has limitations for evaluating complex tables:

\begin{itemize}
    \item It overlooks the positional information of numbers within the table.
    \item Non-numeric content is ignored, which may lead to inaccurate assessments when tables include text.
    \item It fails to distinguish between high and low relative errors, which can result in imprecise evaluations.
    \item RNSS does not account for precision versus recall losses in table reconstruction, limiting its utility for detailed accuracy assessments.
\end{itemize}

\justify
{\textbf{B. Relative Mapping Similarity (RMS) : }} 
\justify
The RMS metric was developed to address the limitations of RNSS by evaluating both numeric and textual content while accounting for the structural layout of the table. RMS interprets tables as a collection of mappings from row and column headers to values, enabling a comparison that respects both value accuracy and positional consistency.

Let , Each entry in the predicted table \( P \) be represented as \( p_i = (p_{r_i}, p_{c_i}, p_{v_i}) \), where \( p_{r_i} \) and \( p_{c_i} \) are row and column headers, and \( p_{v_i} \) is the value. Similarly, each entry in the target table \( T \) be represented as \( t_j = (t_{r_j}, t_{c_j}, t_{v_j}) \).

The RMS metric computes both precision and recall scores based on pairwise distances between corresponding entries in \( P \) and \( T \). The distance for textual entries is calculated with Normalized Levenshtein Distance (NL) \cite{8978179}, and for numeric entries, with the relative distance \( D_{\theta}(p, t) \), defined as:

\[
D_{\theta}(p, t) = \min\left(1, \frac{|p - t|}{|t|}\right)
\]

For each entry pair \( (p_i, t_j) \), the similarity \( D_{\tau, \theta}(p, t) \) is calculated as:

\[
D_{\tau, \theta}(p, t) = (1 - \text{NL}_{\tau}(p_r \| p_c, t_r \| t_c)) \cdot (1 - D_{\theta}(p_v, t_v))
\]

where \( \| \) denotes string concatenation, and the parameter \( \tau \) limits partial credit for highly dissimilar texts.

The RMS Precision and RMS Recall scores are computed by summing similarities over matched entries:

\[
\text{RMS Precision} = 1 - \frac{\sum_{i=1}^{N} \sum_{j=1}^{M} X_{ij} D_{\tau, \theta}(p_i, t_j)}{N}
\]

\[
\text{RMS Recall} = 1 - \frac{\sum_{i=1}^{N} \sum_{j=1}^{M} X_{ij} D_{\tau, \theta}(p_i, t_j)}{M}
\]

The RMS F1 Score, representing overall performance, is computed as the harmonic mean of RMS Precision and RMS Recall:

\[
\text{RMS F1 Score} = 2 \cdot \frac{\text{RMS Precision} \cdot \text{RMS Recall}}{\text{RMS Precision} + \text{RMS Recall}}
\]

This score is invariant to row and column transpositions, making it well-suited for evaluating structured table data.

\subsection*{\textbf{6.2 Evaluation of Base DEPLOT Model on Custom Dataset}}
\justify

Our dataset for testing the base DEPLOT model includes 1,000 bar charts distributed as follows:

\begin{itemize}
    \item 500 Simple Bar Charts
    \item 300 Stacked Bar Charts
    \item 200 Grouped Bar Charts
\end{itemize}

\justify
The RNSS and RMS scores for each chart type are recorded to analyze the model’s performance on both numeric accuracy and structural mapping and the averages were reported as follows for the Base DEPLOT Model:

\begin{itemize}
    \item \textbf{Average RNSS:} 89.67\%
    \item \textbf{Average RMS F1 Score:} 50.93\%
\end{itemize}

\justify
Detailed Results by Chart Type as summarized in Table \ref{tab:table_1}: 

\justify
\textbf{Simple Bar Charts:} Simple bar charts, with one value per category, provide a straightforward test of numeric accuracy. Despite their simplicity, the RMS F1 score (30.62\%) reveals structural alignment challenges, indicating room for improvement in row-column accuracy and value positioning.

\justify
\textbf{Stacked Bar Charts:} Stacked bar charts add complexity by including multiple values per category. The high RNSS score (92.33\%) reflects strong numeric accuracy, while the RMS F1 score (63.34\%) shows that the model faces moderate difficulty in interpreting the structure of stacked values.

\justify
\textbf{Grouped Bar Charts:} Grouped bar charts, which present clusters of bars representing distinct groups, require precise structural mapping. The model achieves high RNSS (98.48\%) and RMS F1 (83.08\%) scores, indicating robust performance in handling complex arrangements and maintaining row-column consistency.

\begin{table}[H]
    \centering
    \renewcommand{\arraystretch}{1.5} 
    \setlength{\tabcolsep}{5pt} 
    \begin{tabular}{|p{0.14\textwidth}|p{0.13\textwidth}|p{0.13\textwidth}|}
    \hline
    \textbf{Chart Type} & \textbf{RNSS Score (\%)} & \textbf{RMS F1 Score (\%)} \\ \hline
    Simple Bar Charts   & 84.55                    & 30.62                      \\ \hline
    Stacked Bar Charts  & 92.33                    & 63.34                      \\ \hline
    Grouped Bar Charts  & 98.48                    & 83.08                      \\ \hline
    \end{tabular}
     \vspace{2mm}
    \caption{PERFORMANCE OF BASE DEPLOT MODEL ON DIFFERENT BAR CHART TYPES}
   
    \label{tab:table_1}
\end{table}

\section*{\textbf{7. Fine-Tuning DEPLOT on the Custom Bar Chart Dataset}}
\justify

Accurate representation of charts as structured tables is crucial for facilitating efficient querying and reasoning, especially when leveraging the zero/few-shot inference capabilities of large language models (LLMs). In particular, a well-defined table structure enhances the model's ability to perform complex reasoning tasks without the need for extensive retraining.

The accurate mapping of chart elements (e.g., axis labels, categories, and values) into tables provides a structured format that is more easily parsed by the LLM, facilitating tasks such as data extraction, trend analysis, and question answering. In turn, this structure empowers users to query the model effectively.

In this section, we describe the process of fine-tuning the DEPLOT model on a specialized dataset of 50,000 financial bar chart images. The dataset is divided as follows: 50\% simple bar charts, 30\% stacked bar charts, and 20\% grouped bar charts. This distribution ensures that the model is exposed to a wide range of chart structures, improving its ability to handle diverse visualization types. This fine-tuning is intended to enhance the model's accuracy and specificity in interpreting bar charts that include financial data. The fine-tuning is conducted using structured data on financial metrics, making the model more proficient in financial terminology, numerical patterns, and chart layouts common in finance.

\subsection*{\textbf{7.1 Dataset Preparation}}
\begin{figure}
    \centering
    \includegraphics[width=1\linewidth]{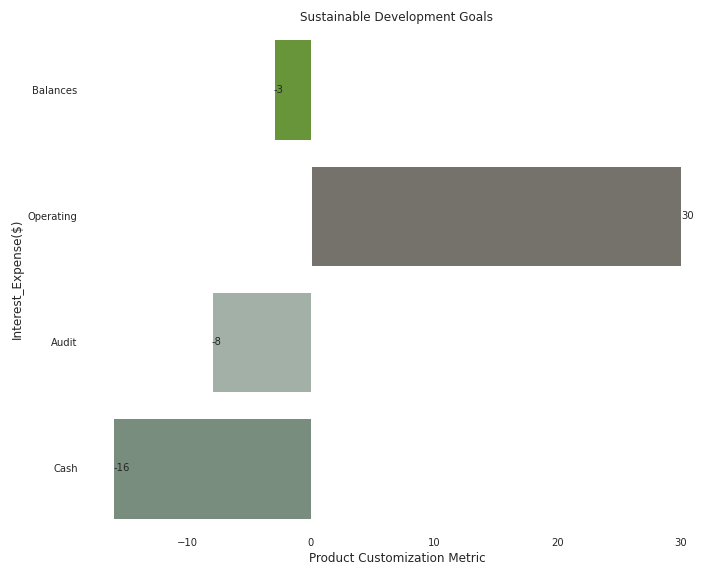}
    \caption{\textit{\textbf{(\(G_t)\):} ``TITLE \textbar Sustainable Development Goals \texttt{<0x0A>} Interest\_Expense(\$) \textbar Product Customization Metric \texttt{<0x0A>} Cash \textbar -16 \texttt{<0x0A>} Audit \textbar -8 \texttt{<0x0A>} Operating \textbar 30 \texttt{<0x0A>} Balances \textbar -3"}}
    \label{fig:enter-label_1}
\end{figure}

\begin{figure}
    \centering
    \includegraphics[width=1\linewidth]{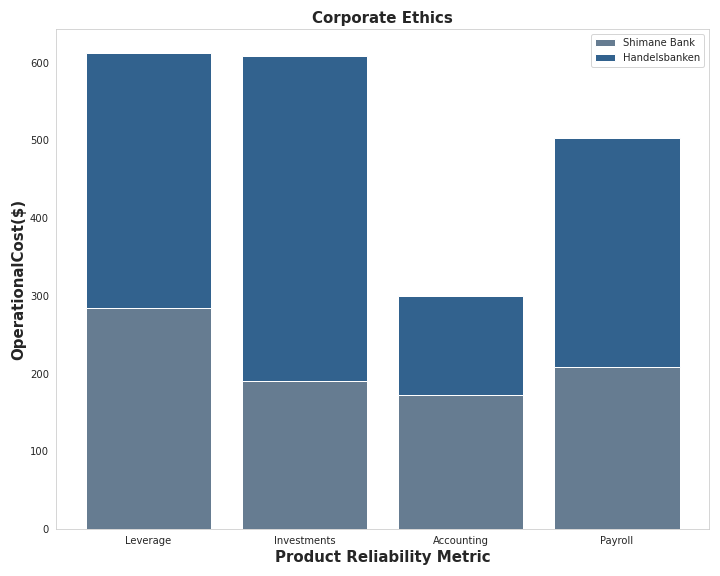}
    \caption{\textit{\textbf{(\(G_t)\):} ``TITLE \textbar Corporate Ethics \texttt{<0x0A>} Product Reliability Metric \textbar Shimane Bank \textbar Handelsbanken \texttt{<0x0A>} Leverage \textbar 285 \textbar 328 \texttt{<0x0A>} Investments \textbar 191 \textbar 418 \texttt{<0x0A>} Accounting \textbar 173 \textbar 127 \texttt{<0x0A>} Payroll \textbar 209 \textbar 294"}}
    \label{fig:enter-label_2}
\end{figure}

\begin{figure}
    \centering
    \includegraphics[width=1\linewidth]{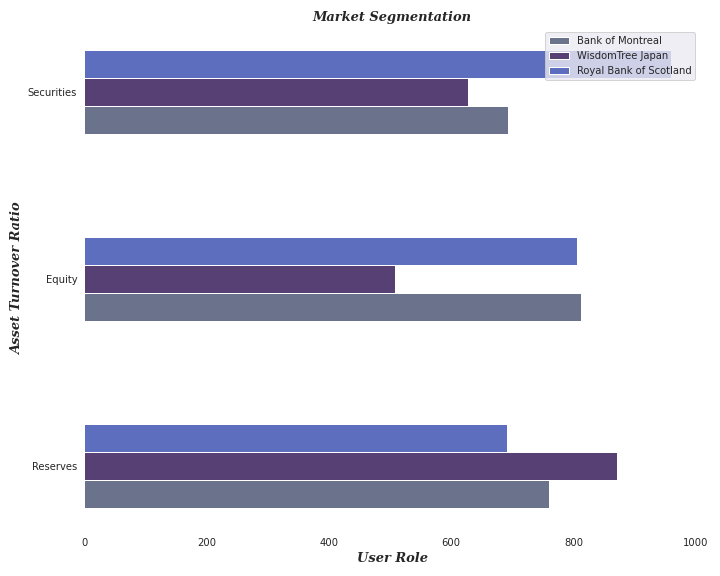}
    \caption{\textit{\textbf{(\(G_t)\):} ``TITLE \textbar Market Segmentation \texttt{<0x0A>} Asset Turnover Ratio \textbar Bank of Montreal \textbar WisdomTree Japan \textbar Royal Bank of Scotland \texttt{<0x0A>} Reserves \textbar 759 \textbar 870 \textbar 691 \texttt{<0x0A>} Equity \textbar 812 \textbar 508 \textbar 805 \texttt{<0x0A>} Securities \textbar 692 \textbar 627 \textbar 959"}}
    \label{fig:enter-label_3}
\end{figure}

\justify
The dataset used for fine-tuning comprises 50,000 image-text pairs, where each image represents a bar chart and the corresponding ground truth table (\(G_t)\) provides a structured syntax representation of the underlying data for the chart. 
Each data point includes (As shown in Fig \ref{fig:enter-label_1}, \ref{fig:enter-label_2}, and \ref{fig:enter-label_3}) :

\begin{itemize}
    \item \textbf{Image Data:} Each bar chart is saved as an image file, representing a visualization of financial data in various formats.
    \item \textbf{Text Data (\(G_t)\):} Each chart has a corresponding text-based data table with proper syntax. The table text includes information like the chart title, labels, and values for each category, mirroring the layout of the bar chart.
\end{itemize}

To prepare this dataset:

\begin{enumerate}
    \item The JSON file containing the ground truth tables is loaded, with each entry corresponding to a specific chart image.
    \item Each entry is structured to include the image of the bar chart and the table data in textual form, creating a dataset that aligns with the input-output structure expected by the DEPLOT model.
\end{enumerate}

\subsection*{\textbf{7.2 Model Configuration and Dataset Processing}}
\justify
To effectively train DEPLOT, the model configuration and data processing are tailored for bar chart interpretation.

\begin{itemize}
    \item \textbf{Model Initialization:} The fine-tuning uses the DEPLOT model pre-trained on general chart data. This model serves as a foundation, and it is then further trained to specialize in financial bar charts through the custom dataset.
    \item \textbf{Text Processing:} The textual data from each table is processed with tokenization and padding, which are essential for consistent input format across different chart types. These steps ensure that the textual ground truth for each image is aligned with the structure that the DEPLOT model expects.
    \item \textbf{Image Preprocessing:} Each bar chart image is processed into a series of image patches to match DEPLOT’s expected input structure. To handle the complexity of detailed financial charts, we configure the image inputs to include up to 1,024 patches per image. This setting allows the model to capture fine-grained visual features, such as text labels and segment borders in the charts.
\end{itemize}

\subsection*{\textbf{7.3 Training Procedure}}
\justify
The fine-tuning process is conducted over multiple epochs with the goal of improving DEPLOT’s performance on financial bar charts.

\begin{itemize}
    \item \textbf{Training Configuration:} The fine-tuning is set to run for 10 epochs, with the model leveraging an AdamW optimizer to update weights effectively and maintain stability in the learning process. The training is conducted on a NVIDIA H100 GPU to speed up the process.
    \item \textbf{Training Loop:} For each epoch, the model processes batches of images and text data from the dataset. In each batch, the model is prompted to generate the data table that corresponds to the input image, effectively ``learning" the relationships between the visual representation and the underlying numerical data. The model computes loss values during this process, which measure the difference between the predicted and actual tables, and then adjusts its weights to minimize these errors.
    \item \textbf{Checkpointing:} To monitor progress and safeguard against any potential interruptions, model weights are saved at the end of each epoch. These checkpoints serve as backup models that represent the state of the model at specific stages of training, allowing for flexibility in further analysis or adjustments.
\end{itemize}

This fine-tuning process aims to adjust the DEPLOT model to recognize the nuanced details of financial bar charts, refining its ability to interpret both numeric values and structural layouts within these charts. By training the model on a diverse set of bar chart examples, we ensure that it becomes proficient in decoding a variety of financial chart types, preparing it for the final evaluation phase of the study.

The fine-tuning process is expected to yield a model with improved capability to interpret and convert complex bar charts into structured tables, which is critical for applications in finance where precise data extraction from charts is essential.

\section*{\textbf{8. Evaluation of Fine-Tuned DEPLOT Model}}
\justify
This section analyzes the improvements observed in the DEPLOT model’s performance after fine-tuning it for 10 epochs on a custom dataset of financial bar charts. We compare the fine-tuned model’s scores with those of the base model to demonstrate how the specialized training data enhanced the model's interpretative accuracy.

\subsection*{\textbf{8.1 Performance Metrics}}
\justify
To assess the improvements rigorously, we employed two primary metrics:

\justify
\textbf{Relative Number Set Similarity (RNSS)}: This metric measures the degree of similarity between the numeric values in the predicted and target tables. It focuses solely on numerical accuracy by comparing the sets of numbers in both tables, irrespective of order or structure.

\justify
\textbf{Relative Mapping Similarity (RMS)}: This metric evaluates both structural and positional similarity between the predicted and target tables. It accounts for both numeric and textual content, computing the accuracy of mapping values to their respective categories or labels. RMS combines precision and recall across mappings of row and column headers to values, providing a more detailed evaluation than RNSS.

\begin{table} [H]
    \centering
    \renewcommand{\arraystretch}{1.5}
    \setlength{\tabcolsep}{5pt} 
    \begin{tabular}{|p{0.1\textwidth}|p{0.15\textwidth}|p{0.15\textwidth}|}
    \hline
    \textbf{Epoch}      & \textbf{RNSS Score (\%)} & \textbf{RMS F1 Score (\%)} \\ \hline
        1       & 98.00                   & 92.89                       \\ \hline
        2       & 97.87                   & 91.51                       \\ \hline
        3       & 98.04                   & 93.32                       \\ \hline
        4       & 97.60                   & 90.03                       \\ \hline
        5       & 97.46                   & 88.42                       \\ \hline
        6       & 98.07                   & 91.06                       \\ \hline
        7       & 97.77                   & 85.71                       \\ \hline
        8       & 97.02                   & 87.49                       \\ \hline
        9       & 96.78                   & 88.53                       \\ \hline
        10      & 96.80                   & 86.37                       \\ 
    \hline
    \end{tabular}
    \vspace{2mm}
    \caption{RMS $F_1$ AND RNSS SCORES FOR EACH MODEL ACROSS EPOCHS DURING FINE-TUNING}
    \label{tab:table_epoch}
\end{table}

Table \ref{tab:table_epoch} demonstrates the progression of the RNSS and RMS F1 scores for the fine-tuned model across ten epochs. Notably, the fine-tuned model achieves its highest RNSS score of 98.07\% and an RMS F1 score of 91.51\% at epoch 6 and epoch 2, respectively, signifying a marked improvement in both mapping similarity and interpretative accuracy. These scores reflect the effectiveness of the fine-tuning process in enhancing the model's ability to extract precise tabular representations from bar charts. \\

\indent In Table \ref{tab:table_second}, the fine-tuned model at epoch 6 is directly compared to the Base DEPLOT model. The base model, with RNSS and RMS F1 scores of 89.67\% and 50.93\%, respectively, exhibits significantly lower performance. This stark contrast highlights the limitations of the base model in handling domain-specific data and the substantial gains achieved through fine-tuning. The RNSS score improvement of approximately 8.4 percentage points and the RMS F1 score improvement of over 40 percentage points underscore the critical role of tailored training on a specialized dataset.\\

\indent In addition to the numerical results provided in Table \ref{tab:table_second}, we further illustrate the model's improvements through visual examples. \\

\begin{figure}[htbp]
    \centering
    \includegraphics[width=1\linewidth]{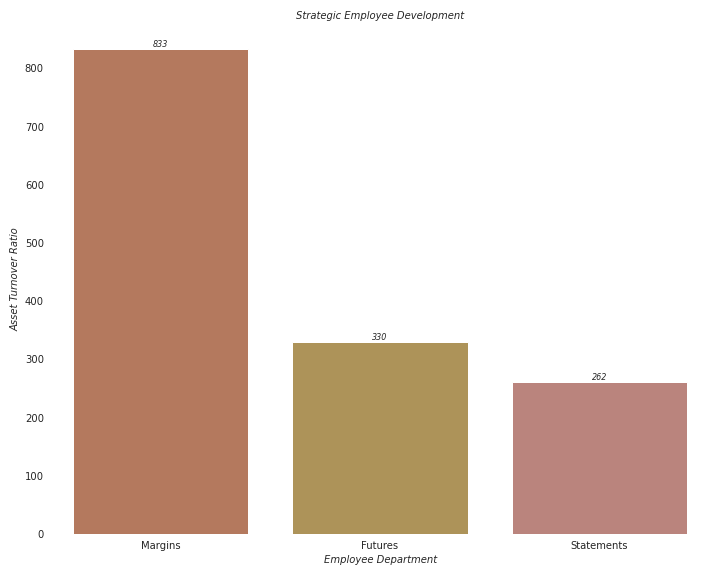}
    \caption{\textit{\textbf{(\(F_t)\)}: TITLE \textbar Strategic Employee Development \texttt{<0x0A>} Employee Department \textbar Asset Turnover Ratio \texttt{<0x0A>} Margins \textbar 833 \textbar Futures \textbar 328 \textbar Statements \textbar 262 \\ 
    \textbf{(\(B_t\))}: TITLE \textbar \texttt{<0x0A>} Strategic Employee Development \textbar Struct E\texttt{<0x0A>} Margins \textbar 83 \textbar Futures\texttt{<0x0A>} Employee Development \textbar 327 \textbar Statements \textbar 258}
    }
    \label{fig:finetuned-vs-base}
\end{figure}

\indent Both the Fine-tuned Model Table (\(F_t\)) and Base Model table (\(B_t\)) are presented for comparison in Fig \ref{fig:finetuned-vs-base}.\\

\indent For Fig \ref{fig:finetuned-vs-base}, The fine-tuned model shows improved accuracy in mapping financial terms and values. The base model outputs misaligned terms such as ``Struct E” and incorrect numerical values (e.g., 83 for ``Margins"), while the fine-tuned model provides the correct terms like ``Asset Turnover Ratio” and accurate values (e.g., 833 for ``Margins"). This demonstrates the fine-tuned model’s enhanced precision in interpreting financial data.

\begin{table}
    \centering
    \renewcommand{\arraystretch}{1.5} 
    \setlength{\tabcolsep}{5pt}
    \begin{tabular}{|p{0.14\textwidth}|p{0.13\textwidth}|p{0.13\textwidth}|}
    \hline
    \textbf{Model} & \textbf{RNSS Score (\%)} & \textbf{RMS F1 Score (\%)} \\
    \hline
    Base DEPLOT Model   & 86.67\%                    & 50.93\%                      \\
    \hline
    Fine-Tuned DEPLOT  & 98.07\%                    & 91.06\%                      \\ 
    \hline   
    \end{tabular}
    \vspace{2mm}
    \caption{COMPARISON OF RNSS AND RMS F1 SCORES BETWEEN THE BASE AND FINE-TUNED DEPLOT MODELS.}
    \label{tab:table_second}
\end{table}

These results collectively validate the hypothesis that fine-tuning DEPLOT on a domain-specific dataset can drastically enhance its interpretative accuracy and relevance for financial bar chart analysis.

\justify
\subsection*{\textbf{8.2 Reasons for Improvement}}

\justify
The observed improvements can be attributed to several factors:

\begin{itemize}
    \item \textbf{Domain-Specific Labeling and Structure}: The base DEPLOT model, being generalized, may not be optimized for interpreting financial terminology or complex layouts commonly seen in financial reports. Fine-tuning a dataset containing financial bar charts with tailored labels and structured terminology allowed the model to better understand and interpret the specific lexicon and structure of financial data.
    \item \textbf{Variety in Chart Types and Configurations}: The custom dataset includes simple, stacked, and grouped bar charts, as well as variations in orientation (horizontal and vertical). By exposing the model to diverse formats within a single domain, fine-tuning helped it learn context-specific patterns and relationships within each chart type. This variety was especially beneficial for RMS scores, as the model learned the mappings between categories, values, and labels more accurately across different structures.
    \item \textbf{Contextual Numerical Range}: In financial contexts, numbers can span several magnitudes, from small percentage changes to large absolute values. By training on data with realistic financial value ranges, the model developed sensitivity to these numerical contexts, allowing it to make more precise numerical predictions in test cases. The improved RNSS score suggests that the model learned to recognize realistic financial data distributions, enabling it to maintain accuracy even with fluctuating value ranges.
    \item \textbf{Reinforcement of Text-Number Associations}: Fine-tuning reinforced the associations between specific text and numeric values. These associations are fundamental in financial reports, where context is critical to interpreting data accurately. The model’s training included multiple scenarios of such associations, allowing it to map numbers to their respective financial terms better, as reflected in the marked increase in RMS scores.
\end{itemize}

\subsection*{\textbf{8.3 Importance of RMS Score in Chart Data Interpretation}}
\justify

While both RNSS and RMS scores are critical for evaluating the model’s accuracy, the RMS score holds particular significance in the context of practical applications. A higher RMS score indicates that the model can accurately map numeric values to their respective row and column headers, effectively reconstructing the table in a structured format. This precise mapping is crucial, especially in financial data interpretation, where even minor misalignments between labels and values can lead to misinterpretations.

Once the table is constructed accurately, it opens up new possibilities for enhanced data interaction. With a well-structured table, we can leverage advanced large language models (LLMs) like GPT-4o or LLaMA for querying and reasoning. These models, with their few-shot learning capabilities, enable sophisticated user querying, allowing users to extract insights or answer complex questions about the data without needing to navigate the underlying table structure directly.

In this context, the RMS score becomes a gateway to making financial data more accessible and interactive. For instance, accurately structured tables can be queried for trends, comparative analysis, or financial projections, all driven by the precise mappings learned through fine-tuning. By prioritizing and achieving a high RMS score, the model enhances its utility beyond basic data extraction, enabling powerful querying capabilities and deeper insights into financial datasets.

\subsection*{\textbf{8.4 Importance of Proper Table Construction for Advanced Reasoning with LLMs}}

\justify
Accurate table construction plays a crucial role in leveraging large language models (LLMs) for advanced reasoning and query answering. Properly structured tables ensure that the information is correctly interpreted, leading to accurate and reliable responses when queried. In financial data extraction, this becomes particularly important, as even minor misalignments between data points and their corresponding labels can result in incorrect interpretations.

For instance (In Fig \ref{fig:finetuned-vs-base}), consider two tables produced by the base and fine-tuned versions of the DEPLOT model. In the base model’s output, the table contains misaligned terms such as ``Struct E" and an incorrect value of 83 for ``Margins," which could confuse an LLM querying the table. When asked, ``What is the value of Margins?”, GPT-4 provides the wrong answer, 83, as it fails to correctly map the financial context.

In contrast, the fine-tuned model produces a correctly structured table, with accurate terms and values, such as ``Asset Turnover Ratio" and 833 for ``Margins." When the same question is posed to GPT-4, it correctly identifies 833 as the value of ``Margins."

This comparison demonstrates how crucial precise table construction is for effective reasoning. When the model outputs a well-structured table, LLMs can use their few-shot learning capabilities to understand the context better and generate accurate answers. By ensuring the underlying table is constructed properly, advanced reasoning tasks like trend analysis, financial comparisons, and predictive insights become more reliable and actionable, thereby enhancing the utility of data processing systems.

\section*{\textbf{9. Improving Question Answering: A Table-Driven Approach}}
\justify

Recent advances in Large Language Models (LLMs) have enabled promising capabilities for multimodal reasoning and question answering (QA) over visual inputs such as charts and plots. However, directly providing a plot image to a VLM often results in a suboptimal quantitative understanding of the underlying data distribution. To address this gap, we propose a two-step approach: first, employ a state-of-the-art table-extraction model, such as DEPLOT, to infer the underlying data table from the input chart, and second, query the LLM using the extracted tabular data rather than the raw image. Additionally, we refine this method by fine-tuning the DEPLOT model to adapt to specific chart characteristics in the finance domain. Empirical results demonstrate significant performance improvements in QA accuracy, as measured by classical error metrics, and even enable smaller vision-language models \cite{unknown} \cite{phogat2024finetuningsmallerlanguagemodels} to surpass more capable but less specialized models.

\subsection*{\textbf{9.1 Experimental Setup}}
\justify

\textbf{LLM Pool and Configurations:} We consider a pool of LLMs with varying capacities and training regimes: 
\[
\mathcal{M} = \{\text{GPT-4o}, \text{Qwen-2b-VL}, \text{Qwen-7b-VL},
 \text{LLaMa-11B}\}.
\]

For each model $m \in \mathcal{M}$, we evaluate three configurations:
\begin{enumerate}
    \item \textbf{Image-Only:} The model is provided only with the plot image (Img) directly.
    \item \textbf{Base-Table:} The model is provided with a table ($B_t$) generated by the base DEPLOT model and the image.
    \item \textbf{Fine-Tuned Table:} The model is provided with a table ($F_t$) generated by the fine-tuned DEPLOT model that has been adapted to the target domain.
\end{enumerate}

\justify
\textbf{Data and Questions:} We select 100 finance-domain charts (simple, stacked, and grouped barcharts) for evaluation. Each chart is presented to the LLM in one of the three configurations mentioned above, and the LLM generates answer candidates for a carefully curated set of question-answer pairs. The questions and answers were generated using GPT-4o and verified manually to ensure correctness and diversity.

\justify
\textbf{Metrics:} To quantify the accuracy of the numeric reasoning performed by the LLMs, we employ classical predictive error metrics. 
\justify
\indent Let $\{y_i\}_{i=1}^N$ be the ground-truth numeric values (e.g., data points from the chart) and $\{\hat{y}_i\}_{i=1}^N$ be the values predicted by the LLM. We consider the following.

\begin{enumerate}
    \item \textbf{Mean Absolute Percentage Error (MAPE):}
    \[
    \text{MAPE} = \frac{100\%}{N} \sum_{i=1}^N \frac{|y_i - \hat{y}_i|}{|y_i|}.
    \]
    MAPE measures relative errors, offering a scale-independent evaluation of performance.
    \item \textbf{Root Mean Squared Error (RMSE):}
    \[
    \text{RMSE} = \sqrt{\frac{1}{N} \sum_{i=1}^N (y_i - \hat{y}_i)^2}.
    \]
    RMSE emphasizes larger errors due to squaring and is sensitive to outliers.
\end{enumerate}

\justify
By simultaneously examining MAPE and RMSE, we gain a nuanced view of both relative and absolute performance across different LLM and configuration settings.

\subsection*{\textbf{9.2 Result Analysis}}

\justify
In Table \ref{tab:results_metrics}, $\text{MAPE}(m_c)$ and $\text{RMSE}(m_c)$ represent the MAPE and RMSE, respectively, for model $m$ under configuration $c$.

\begin{table}[h!]
\centering
\renewcommand{\arraystretch}{1.3} 
\setlength{\tabcolsep}{8pt}       
\begin{tabular}{l c c}
\toprule
\textbf{Configuration, $c$}            & \textbf{MAPE (\%)} & \textbf{RMSE (\%)} \\ \midrule
\multicolumn{3}{l}{\textbf{GPT-4o}} \\
\hspace{2mm} + Img  + Q                      & 16.07              & 60.51             \\
\hspace{2mm} + Img + $B_t$ + Q               & 11.09              & 65.89             \\
\hspace{2mm} + Img + $F_t$ + Q               & \textbf{3.66}      & \textbf{17.01}    \\ \midrule
\multicolumn{3}{l}{\textbf{Qwen-2b}} \\
\hspace{2mm} + Img + Q                      & 37.81              & 80.84             \\
\hspace{2mm} + Img + $B_t$  + Q               & 32.84              & 90.52             \\
\hspace{2mm} + Img + $F_t$  + Q               & 23.04              & 59.77             \\ \midrule
\multicolumn{3}{l}{\textbf{Qwen-7b}} \\
\hspace{2mm} + Img   + Q                     & 32.63              & 90.99             \\
\hspace{2mm} + Img + $B_t$  + Q               & 14.32              & 69.43             \\
\hspace{2mm} + Img + $F_t$  + Q               & \textbf{2.98}      & \textbf{8.10}     \\ \midrule
\multicolumn{3}{l}{\textbf{LLaMa-11B}} \\
\hspace{2mm} + Img    + Q                   & 20.06              & 67.39             \\
\hspace{2mm} + Img + $B_t$ + Q                & 9.86               & 65.13             \\
\hspace{2mm} + Img + $F_t$  + Q               & \textbf{2.06}      & \textbf{9.00}     \\ \bottomrule
\end{tabular}
\vspace{3mm}
\caption{AVERAGE $\text{MAPE}(m_c)$ AND $\text{RMSE}(m_c)$ FOR EACH MODEL $m$ AND CONFIGURATION $c$. HERE, \textbf{$B_t$}REPRESENTS THE BASE TABLE, AND \textbf{$F_t$} REPRESENTS THE FINE-TUNED TABLE.}
\label{tab:results_metrics}
\end{table}

\justify
\textbf{Effectiveness of Table Extraction}
\justify
For all models ($m \in \mathcal{M}$), transitioning from the image-only configuration to the fine-tuned table configuration consistently and significantly reduces both Mean Absolute Percentage Error (MAPE) and Root Mean Square Error (RMSE).

\justify  
Formally, let:  
\[
\Delta \text{MAPE}^c_m = \left| \text{MAPE}^{c_1}_m - \text{MAPE}^{c_2}_m \right|
\]  
and similarly for RMSE:  
\[
\Delta \text{RMSE}^c_m = \left| \text{RMSE}^{c_1}_m - \text{RMSE}^{c_2}_m \right|.
\]  

Here, \(c_1\) and \(c_2\) represent any two configurations being compared (e.g., Image-Only, Base-Table, or Fine-Tuned Table), while \(m\) remains the same model.

\justify
For GPT-4o, the improvement in MAPE is calculated as follows:
\[
\Delta \text{MAPE}_{\text{GPT-4o}} = 16.07 - 3.66 = 12.41\%,
\]
which corresponds to a 77.2\% reduction in relative terms.

\justify
For Qwen-7b, the improvement is even more substantial:
\[
\Delta \text{MAPE}_{\text{Qwen-7b}} = 32.63 - 2.98 = 29.65\%,
\]
resulting in an approximate 90.9\% reduction in relative terms.

\justify
These notable improvements highlight the pivotal role of precise tabular data extraction in enhancing the accuracy of numerical reasoning tasks.
\justify
\textbf{Comparison Among Models}
\justify

An intriguing result is that Qwen-7b and LLaMa-11B, when provided with finetuned tables, outperform GPT-4o on both MAPE and RMSE metrics. For example:
\[
\text{GPT-4o: MAPE} = 3.66\%, \quad \text{RMSE} = 17.01\%,
\]
while LLaMa-11B achieves:
\[
\text{MAPE} = 2.06\%, \quad \text{RMSE} = 9.00\%.
\]
This result suggests that when data is presented in a clearer, more structured form (i.e., tabular format), smaller models with fewer parameters can outperform larger, more capable models that must visually parse the raw data.

To quantify this difference, we define a relative improvement metric $\rho$ for MAPE between two models $m_1$ and $m_2$ as:
\[
\rho(m_1, m_2) = \frac{\text{MAPE}(m_1) - \text{MAPE}(m_2)}{\text{MAPE}(m_1)} \times 100\%.
\]
For the finetuned scenario between GPT-4o and LLaMa-11B:
\[
\rho(\text{GPT-4o}, \text{LLaMa-11B}) = \frac{3.66 - 2.06}{3.66} \approx 43.7\% 
\]

\justify
\textbf{Impact of Fine-Tuning the DEPLOT Model}
\justify

The comparison between the base-table and fine-tuned table configurations highlights the significant impact of fine-tuning. The fine-tuned DEPLOT model produces tables that more accurately represent the underlying chart data, effectively reducing systemic extraction errors. 

\justify
For the Qwen-2b model:
\[
\begin{aligned}
\text{MAPE}(qwen_{\text{base\_table}}) &= 32.84\%, \\
\text{MAPE}(qwen_{\text{finetuned\_table}}) &= 23.04\%.
\end{aligned}
\]
The relative reduction in MAPE achieved through fine-tuning is:
\[
\frac{32.84 - 23.04}{32.84} \approx 29.8\%.
\]

Similar improvements are observed across all models, illustrating that domain-adaptive fine-tuning directly enhances downstream question-answering (QA) performance by improving the quality of table extraction.

\justify
\textbf{Consistency Across Metrics}
\justify

The improvements are evident not only in MAPE but also in RMSE. Since RMSE penalizes large outliers more heavily than MAPE, the significant drop in RMSE for the fine-tuned table configurations highlights the ability of the fine-tuned DEPLOT model to mitigate catastrophic extraction errors. 

For the Qwen-7b model, RMSE decreases from 90.99\% in the image-only scenario to just 8.10\% in the fine-tuned scenario, representing a reduction by a factor of more than 11. This demonstrates that the model's final predictions are substantially closer to the ground truth, with far fewer extreme deviations.

\justify
\textbf{Generalization to Varied Chart Types}
\justify

Our dataset includes a diverse range of chart types, such as simple, stacked, and grouped bar charts. The consistent improvements observed across all chart types confirm the generalizability of the approach. The fine-tuned DEPLOT model effectively learns robust, domain-specific representations that accurately translate visual encodings (e.g., bar heights and stacked segments) into textual or numerical formats. 

Once structured input is provided, large language models (LLMs), originally trained on textual corpora, leverage this structured data to enhance internal reasoning and generate more accurate numerical outputs.

\section*{\textbf{7. Conclusion and Future Directions}}
\justify

Our two-step approach—extracting tabular data from charts followed by querying LLMs with these tables—demonstrates significant improvements in numeric question-answering (QA) performance. The fine-tuning of the DEPLOT model played a pivotal role, yielding tabular data that substantially reduced MAPE and RMSE across all tested models. Notably, smaller vision-language models, when provided with refined tabular representations, outperformed larger models (such as GPT-4o) operating directly on raw images. This highlights the potential of carefully designed intermediate structured representations to enhance both performance and resource efficiency in complex reasoning tasks.

\justify
Future research could explore several avenues to further advance this approach:
\begin{itemize}
    \item \textbf{Adaptive Data Augmentation}: Leveraging synthetic charts to expand fine-tuning datasets, improving DEPLOT’s generalization to unseen data.
    \item \textbf{Model Ensembles}: Combining multiple table extraction models or integrating error-correction mechanisms to refine extracted tables further.
    \item \textbf{Multimodal Fusion Architectures}: Developing architectures that incorporate refined tabular data alongside selective visual cues for enhanced numeric reasoning.
\end{itemize}

Despite its promising results, the proposed approach has certain limitations. It relies heavily on high-quality, annotated training datasets, making it susceptible to biases or inaccuracies in the data. The model has shown strong performance on simple, stacked, and grouped bar charts but has not been thoroughly evaluated on more complex visualizations (e.g., multi-axis plots or charts with intricate legends). Furthermore, the approach assumes clean input images and consistent visual encodings; any noise or OCR errors may degrade performance. Lastly, domain-specific fine-tuning may limit generalization to unfamiliar chart types or unconventional encodings. Addressing these challenges in future work will enhance the robustness and scalability of the approach.

\section*{\textbf{Acknowledgments}}
\justify
This research was conducted at Synechron Innovation Labs, Bangalore. We would like to extend our heartfelt gratitude to Hareesha Pattaje, Managing Director - Technology, Synechron, for funding and supporting this work. His contributions were instrumental in shaping the direction and success of this research.

\clearpage
\newpage 
\bibliographystyle{IEEEtran}
\bibliography{references}

\clearpage
\newpage  

\textbf{A. COMPARATIVE INFERENCE STUDY}

\begin{figure}[H]
    \centering
    \includegraphics[width=1\linewidth]{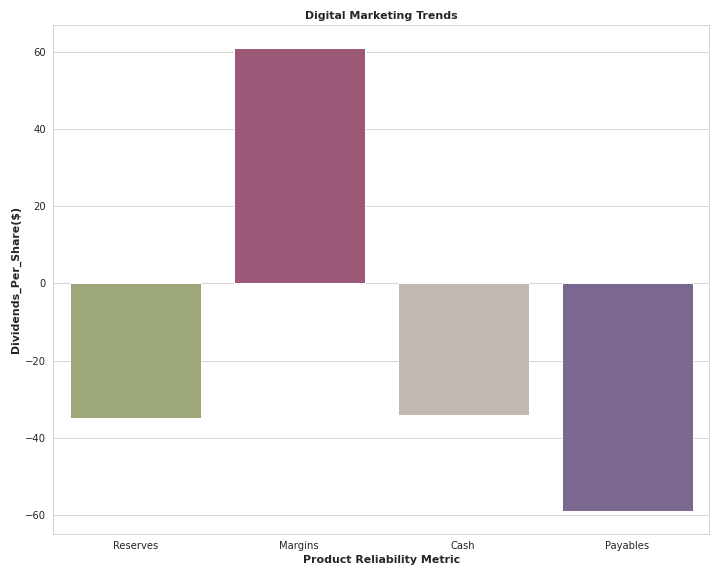}
    \caption{\textbf{Ground Truth (\(G_t\))} :\textit{``TITLE \textbar Digital Marketing Trends \texttt{<0x0A>} Product Reliability Metric \textbar Dividends\_Per\_Share(\$) \texttt{<0x0A>} Reserves \textbar -35 \textbar Margins \textbar 61 \textbar Cash \textbar -34 \textbar Payables \textbar -59''}}

    \label{fig:simple_case}
\end{figure}

\begin{figure}[H]
    \centering
    \includegraphics[width=1\linewidth]{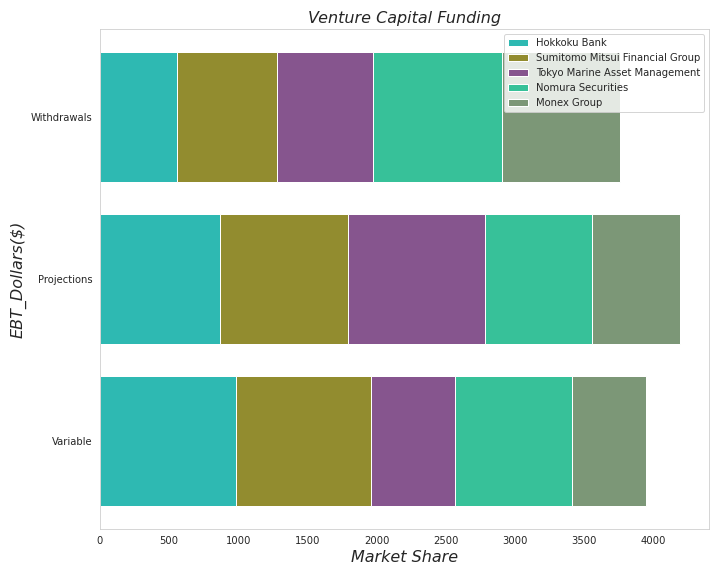}
    \caption{\textbf{Ground Truth (\(G_t\))} :\textit{``TITLE \textbar Venture Capital Funding \texttt{<0x0A>} EBT\_Dollars(\$) \textbar Hokkoku Bank \textbar Sumitomo Mitsui Financial Group \textbar Tokyo Marine Asset Management \textbar Nomura Securities \textbar Monex Group \texttt{<0x0A>} Variable \textbar 985 \textbar 978 \textbar 605 \textbar 840 \textbar 541 \texttt{<0x0A>} Projections \textbar 870 \textbar 922 \textbar 994 \textbar 767 \textbar 640 \texttt{<0x0A>} Withdrawals \textbar 560 \textbar 717 \textbar 699 \textbar 927 \textbar 852''}
}
    \label{fig:stacked_case}
\end{figure}

\begin{figure}[H]
    \centering
    \includegraphics[width=1\linewidth]{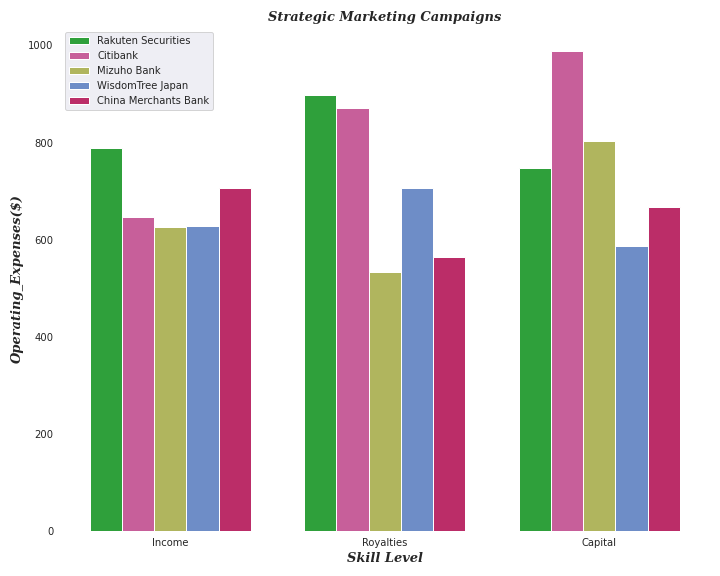}
    \caption{\textbf{Ground Truth (\(G_t\))} :\textit{``TITLE \textbar Strategic Marketing Campaigns \texttt{<0x0A>} Skill Level \textbar Rakuten Securities \textbar Citibank \textbar Mizuho Bank \textbar WisdomTree Japan \textbar China Merchants Bank \texttt{<0x0A>} Income \textbar 788 \textbar 647 \textbar 627 \textbar 629 \textbar 706 \texttt{<0x0A>} Royalties \textbar 898 \textbar 871 \textbar 533 \textbar 707 \textbar 565 \texttt{<0x0A>} Capital \textbar 748 \textbar 989 \textbar 804 \textbar 588 \textbar 667''}}
    \label{fig:grouped_case}
\end{figure}

\vspace{12pt}  
\textbf{A.1 Base Model (\(B_t\)) vs. Fine-tuned Model (\(F_t\))}  
\justify  
This section provides a comparative analysis of the table-generation performance of the Base Model (\(B_t\)) and the Fine-tuned Model (\(F_t\)). The evaluation utilizes three representative figures from the test dataset (Fig. \ref{fig:simple_case}, \ref{fig:stacked_case}, \ref{fig:grouped_case}). Table \ref{tab:comparison} illustrates the tables produced by each model for these figures, with the row separator ``\texttt{<0x0A>}'' replaced by line breaks to enhance readability. This comparison assesses the impact of fine-tuning on the DEPLOT model's accuracy and reasoning capabilities.  

\vspace{12pt}  

\textbf{A.2 Zero-Shot Inference: Intermediate Tables vs. Direct Chart Queries}  
\justify  
Table \ref{tab:DEPLOT_table_improvements} presents a comparative analysis across three scenarios involving zero-shot inference with the intermediate tables generated (\(G_t\), \(F_t\), \(B_t\)) for querying charts indirectly. For comparison, results obtained by directly querying GPT-4o with the chart images are also included. This study underscores the significance of accurate table generation and its role in effective chart interpretation and mapping.  

\vspace{12pt}  

\textbf{A.3 Vision-Language Models: Zero-Shot Inference with Intermediate Tables vs. Direct Chart Queries}  
\justify  
In Table \ref{tab:SLMSvsDEPLOT}, we compare the performance of advanced vision-language models (VLMs) in interpreting the same three figures used in our evaluation. These models were tasked with directly analyzing the figures without generating intermediate tables. This comparison highlights the relative effectiveness of bypassing intermediate representations versus leveraging them for inference tasks.  

\begin{table*}[ht]
    \centering
    \renewcommand{\arraystretch}{1.5} 
    \setlength{\tabcolsep}{8pt} 
    
    \begin{tabular}{|p{0.05\textwidth}|p{0.425\textwidth}|p{0.425\textwidth}|}
    \hline
    \textbf{FIGURE} & \textbf{BASE MODEL TABLE  (\(G_t)\)} & \textbf{FINE-TUNED MODEL TABLE  (\(F_t\))} \\ \hline
    
    Fig \ref{fig:simple_case} & 
    \texttt{\textbf{TITLE} |Digital Marketing Trends \newline Digital Marketing Trends |P2-Cmerredi \newline Digital Marketing Trends \newline \textbf{\textit{Reserves}} |0 \newline \textbf{\textit{Margius}} \newline \textbf{\textit{Product Reliability}} |61 \newline \textbf{\textit{Dash}} |0 \newline \textbf{\textit{Payables}} |0} &
    \texttt{\textbf{TITLE} |Digital Marketing Trends \newline Product Reliability Metric |Dividends\_Per\_Share(\$) \newline \textbf{\textit{Reserves}} |-35 \newline \textbf{\textit{Margins}} |61 \newline \textbf{\textit{Cash}} |-34 \newline \textbf{\textit{Payables}} |-59} \\ \hline
    
    Fig \ref{fig:stacked_case} & 
    \texttt{\textbf{TITLE} |Venture Capital Funding \newline EBIT |Hokkoku Bank |Sumitomo Mitsui Financial Group |Tokyo Marine Asset Management |Nomura Securities |Money Group \newline \textbf{\textit{Vertical}} |1000 |95.0 |52.0 |82.0 |41.0 \newline \textbf{\textit{Projection}} |880 |880 |885 |75.0 |54.0 \newline \textbf{\textit{Withdrawals}} |540 |630 |630 |75.0 |630} &
    \texttt{\textbf{TITLE} |Venture Capital Funding \newline EBT\_Dollars(\$) | Hokkoku Bank | Sumitomo Mitsui Financial Group | Tokyo Marine Asset Management | Nomura Securities | Monex Group \newline \textbf{\textit{Variable}} |976 |998 |630 |740 |550 \newline \textbf{\textit{Projections}} |860 |940 |998 |760 |670 \newline \textbf{\textit{Withdrawals}} |552 |740 |720 |940 |825} \\ \hline
    
    Fig \ref{fig:grouped_case} & 
    \texttt{\textbf{TITLE} |Strategic Marketing Campaigns \newline Strategic Marketing Campaigns | Rakuten Securities | Citibank | Mizuho Bank | WisdomTree Japan | China Merchants Bank \newline \textbf{\textit{Income}} |785 |645 |625 |626 |705 \newline \textit{\textbf{Royalties}} |897 |868 |530 |705 |562 \newline \textbf{\textit{Digital}} |746 |984 |800 |584 |665} &
    \texttt{\textbf{TITLE} |Strategic Marketing Campaigns \newline Skill Level | Rakuten Securities | Citibank | Mizuho Bank | WisdomTree Japan | China Merchants Bank \newline \textbf{\textit{Income}} |787 |649 |628 |629 |709 \newline \textbf{\textit{Royalties}} |899 |871 |535 |707 |563 \newline \textbf{\textit{Capital}} |746 |986 |804 |586 |669} \\ \hline
    
    \end{tabular}
    ~
    \caption{COMPARISON OF BASE MODEL AND FINE-TUNED MODEL TABLES}
    \label{tab:comparison}
\end{table*}

\vspace{12pt}

\begin{table*}[ht]
    \centering
    \renewcommand{\arraystretch}{1.5} 
    \setlength{\tabcolsep}{6pt} 
    
    \begin{tabular}{|p{0.15\textwidth}|p{0.25\textwidth}|p{0.25\textwidth}|p{0.25\textwidth}|}
    \hline
    \textbf{MODEL} & \textit{\textbf{What is the Value of Reserves and Cash? Also, state the difference in Value.}} [\ref{fig:simple_case}] & 
   \textbf{\textit{What is the value of Projections and Withdrawals for Hokkoku Bank and Nomura Securities?}} [\ref{fig:stacked_case}] & 
    \textit{\textbf{What is the difference between the Capital Value of Citibank and Mizuho Bank?[\ref{fig:grouped_case}}}]
    
 \\ \hline 
    \textbf{(\(G_t\)) + GPT-4o}& The Value of Reserves and Cash are -35 and -34, respectively. The difference in Value is 1. & For Hokkoku Bank, the value of Projections is 870, and Withdrawals is 560.
 For Nomura Securities, the value of Projections is 767, and Withdrawals is 927. & The difference between the Capital value of Citibank (989) and Mizuho Bank (804) is 185. \\ \hline
 
   \textbf{(\(F_t\)) + GPT-4o} & The value of Reserves is -35, the value of Cash is -34, and the difference in value is 1. &  For Hokkoku Bank, the value of Projections is 860, and Withdrawals is 552.
 For Nomura Securities, the value of Projections is 760, and Withdrawals is 940. & The difference between the Capital value of Citibank (986) and Mizuho Bank (804) is 182.  \\ \hline

    \textbf{(\(B_t\)) + GPT-4o} & The value of Reserves is 0, and Cash (interpreted as Dash) is also 0. The difference is 0. & The Projections and Withdrawals for Hokkoku Bank are 880 and 540, respectively, while for Nomura Securities, they are 75.0 and 75.0, respectively. & The text provided does not explicitly mention ``Capital Value" for Citibank or Mizuho Bank. However, it lists values for Income, Royalties, and Digital for both banks. \\ \hline
       \textbf{GPT-4o} & From the provided image, I can extract the approximate values:
    Reserves: Around -20, Cash: Around -40, So the difference in value is 20. & For Hokkoku Bank (cyan color):
Projections: Approximately 1000 (Market Share)
Withdrawals: Approximately 1000 (Market Share)
For Nomura Securities (purple color):
Projections: Approximately 3000 (Market Share)
Withdrawals: Approximately 3000 (Market Share)
Both categories have the same market share values for these entities. & 
From the chart:

Capital value of Citibank (yellow color): Approximately 700 (Operating Expenses).
Capital value of Mizuho Bank (blue color): Approximately 400 (Operating Expenses).
So, the difference between the Capital value of Citibank and Mizuho Bank is 300. \\ \hline 
    
    \end{tabular}
    ~
    \caption{ZERO-SHOT INFERENCE: INTERMEDIATE TABLES VS. DIRECT CHART QUERIES}
    \label{tab:DEPLOT_table_improvements}
\end{table*}

\vspace{12pt}

\begin{table*}[htp]
    \centering
    \renewcommand{\arraystretch}{1.5} 
    \setlength{\tabcolsep}{6pt} 
    
    \begin{tabular}{|p{0.15\textwidth}|p{0.25\textwidth}|p{0.25\textwidth}|p{0.25\textwidth}|}
    \hline
    \textbf{MODEL} & \textit{\textbf{What is the Value of Reserves and Cash? Also, state the difference in Value.}} [\ref{fig:simple_case}] & 
   \textbf{\textit{What is the value of Projections and Withdrawals for Hokkoku Bank and Nomura Securities?}} [\ref{fig:stacked_case}] & 
    \textit{\textbf{What is the difference between the Capital Value of Citibank and Mizuho Bank?[\ref{fig:grouped_case}}}]
    
    \\ \hline 

    \textbf{Qwen2-VL-2B-Instruct} & This bar chart uses shades of gray (possibly green and beige) to represent the 4 different "Product Reliability Metrics": Reserves,...This n-gram (set of 3 n-1 word sequences) for the sample text "Product Reliability Metrics: Reserves,... & To determine the total value of Boca's 1-day spread with the majority of the 1-day and daily distance to the calendar spread and second gap in unordered, we first need to list the 1-day spread and associated distances as follows:| Interval | Traj Value|M1 Value|M2 Value|M3 Value|M4 Value|M5 Value|M6 Value|M7 Val |M1 M2 M3 M4 M5 M6 M7 M8 M9 M10|| & The Capital Value of Mizuho Bank ineson of the two charts under consideration shares a similar presentation with 950 in the y-axes as reported previously  \\ \hline

    \textbf{Phi-3-vision-128k-instruct} \cite{abdin2024phi3technicalreporthighly} & The Value of Reserves is -20 and the Value of Cash is -40. The difference in Value is 20. & The chart shows that Hokkoku Bank has a Projections value of approximately 1500 and a Withdrawals value of approximately 500. Nomura Securities has a Projections value of approximately 2000 and a Withdrawals value of approximately 1000. & The Capital Value of Citibank is approximately \$800 million, and for Mizuho Bank, it is around \$600 million. The difference between their Capital Values is \$800 million - \$600 million = \$200 million.  \\ \hline
    
    \textbf{Qwen2-VL-7B-Instruct} \cite{wang2024qwen2vlenhancingvisionlanguagemodels} & The value of Reserves is -40, and the value of Cash is -20. The difference in value between Reserves and Cash is -40 (Reserves) - (-20) (Cash) = -20. So, the difference in value between Reserves and Cash is 20. & The value of Projections and Withdrawals for Hokkoku Bank and Nomura Securities is 3000. & The difference between the Capital Value of Citibank and Mizuho Bank is 170. \\ \hline
    
    \textbf{Llama-3.2-11B-Vision-Instruct} \cite{touvron2023llamaopenefficientfoundation} & The value of Reserves and Cash is -26 and 0, respectively, and the difference in value is 26. & Looking at the bar plot, we see that the values for Hokkoku Bank and Nomura Securities are as follows:
Hokkoku Bank:
Projections: 1100
Withdrawals: 5300
Nomura Securities:
Projections: 1100
Withdrawals: 5300

  & Looking at the bar plot, I can see the Capital bars for Citibank and Mizuho Bank. Citibank's Capital bar is at 1000, and Mizuho Bank's Capital bar is at 800. To find the difference, I'll subtract Mizuho Bank's Capital from Citibank's Capital.1000 (Citibank's Capital) - 800 (Mizuho Bank's Capital) = 200 \\ \hline
    
    \textbf{Molmo-7B-D-0924} \cite{deitke2024molmopixmoopenweights} & The Value of Reserves is -30 and the Value of Cash is -20. The difference in value between Reserves and Cash is:
    -30 - (-20) = -10
    Therefore, the difference in Value between Reserves and Cash is -10.
 &  For Hokkoku Bank, the values are:

Projections: 1200 million
Withdrawals: 1000 million

For Nomura Securities, the values are:

Projections: 1300 million
Withdrawals: 1100 million
 & The difference between the Capital Value of Citibank and Mizuho Bank is 150. Citibank's Capital Value is 950, while Mizuho Bank's Capital Value is 800. 
 \\ \hline
 
    \textbf{Pixtral-12B-2409 } \cite{agrawal2024pixtral12b} & The value of Reserves is approximately -30, and the value of Cash is approximately -35. So, the difference in value between Reserves and Cash is 5. & Hokkoku Bank, Projections: 1000, Withdrawals: 1000
Nomura Securities, Projections: 2500, Withdrawals: 2500
  & From the chart, the Capital Value for Citibank is approximately 1000, and for Mizuho Bank, it is approximately 800.So, the difference in Capital Value between Citibank and Mizuho Bank is 200. \\ \hline

    \textbf{Phi-3-vision-128k-instruct} \cite{abdin2024phi} & The Value of Reserves is -20 and the Value of Cash is -40. The difference in Value is 20. & The chart shows that Hokkoku Bank has a Projections value of approximately 1500 and a Withdrawals value of approximately 500. Nomura Securities has a Projections value of approximately 2000 and a Withdrawals value of approximately 1000. & The Capital Value of Citibank is approximately \$800 million, and for Mizuho Bank, it is around \$600 million. The difference between their Capital Values is \$800 million - \$600 million = \$200 million.  \\ \hline
    
    \end{tabular}
    ~
    \caption{COMPARISON OF VISION-LANGUAGE MODELS ON INTERPRETING FIGURES}
    \label{tab:SLMSvsDEPLOT}
\end{table*}

\vspace{12pt}

\onecolumn
\clearpage

\textbf{B. PROMPTS}
\vspace{12pt}

This section contains the detailed prompts that were used to generate queries for 100 images, by prompting GPT-4o. Additionally, the prompts used to query the language model (LLM) are included, offering insight into the specific descriptions and instructions given to the model in order to generate Question-answer pairs. 

\vspace{12pt}

\begin{lstlisting}
# Query Generation task prompt

system_prompt = "You are a helpful assistant. Help me with my math homework!"

prompt = f"""
You are an AI assistant tasked with analyzing tabular data extracted from bar chart visualizations. Your job is to generate a single query based on the given table and provide the correct answer as a single integer derived from the data.

Output Requirements:
- Generate one meaningful query based on the table.
- Ensure the query is clear, concise, and specific to a value from the table.
- Provide the correct answer to the query as a single integer.

The following table is represented as text. "|" separates columns (labels and values), and newline marks the end of each row.

{formatted_text}

Provide the output in the following JSON format:
{{
  "query": "<generated question>",
  "correct_answer": <integer answer>
}}
"""
\end{lstlisting}

\vspace{12pt}

\begin{lstlisting}
# Prompt used to query LLMs

system_prompt = "You are a helpful assistant that answers queries based on charts and tables. "
        "Study the inputs and return only the final integer answer. Do not include explanations."
    

prompt = """Study the chart or the given table and think step by step to arrive at the final integer answer. Only return the final integer answer, no description is required.\n"""

\end{lstlisting}
\end{document}